\documentclass[letterpaper]{article}
\usepackage{aaai24}
\usepackage{times} % DO NOT CHANGE THIS
\usepackage{helvet} % DO NOT CHANGE THIS
\usepackage{courier} % DO NOT CHANGE THIS
\usepackage[hyphens]{url} % DO NOT CHANGE THIS
\usepackage{graphicx} % DO NOT CHANGE THIS
\usepackage{graphicx} % DO NOT CHANGE THIS
\usepackage{natbib} % DO NOT CHANGE THIS
\usepackage{caption} % DO NOT CHANGE THIS
\usepackage[utf8]{inputenc} % allow utf-8 input
\usepackage[T1]{fontenc}    % use 8-bit T1 fonts
\usepackage{url}            % simple URL typesetting
\usepackage{booktabs}       % professional-quality tables
\usepackage{amsfonts}       % blackboard math symbols
\usepackage{nicefrac}       % compact symbols for 1/2, etc.
\usepackage{microtype}      % microtypography
\usepackage{xcolor}         % colors
\usepackage{graphicx}

\usepackage{amsmath}
\usepackage{amsthm}
\usepackage{booktabs}
\usepackage{tabularx}
\usepackage{makecell}
\usepackage{algorithm}
\usepackage{multirow}
\usepackage{booktabs}
\usepackage{pdfpages}
\usepackage{graphicx}
\usepackage{subfigure}
\usepackage{bbding}
\usepackage{multibib}
\title{MERGE: Fast Private Text Generation}

\author {
  Zi Liang,  
Pinghui Wang,
   Ruofei Zhang,
    Nuo Xu,
Lifeng Xing, 
Shuo Zhang,
  \\
}
\affiliations {
     MOE KLINNS Lab, Xi'an Jiaotong University, Xi'an 710049, P. R. China \\
    \{liangzid,zs412082986, xlf20200926\}@stu.xjtu.edu.cn, phwang@mail.xjtu.edu.cn,\\ rfzhang@gmail.com, nxu@sei.xjtu.edu.cn
}

\begin{document}

\maketitle

\begin{abstract}
%Recent years have seen increasing concerns about the private inference of NLP services and Transformer models. However, existing two-party privacy-preserving methods solely consider NLU scenarios, while the private inference of text generation such as translation, dialogue, and code completion remains unsolved. Besides, while migrated to NLG models, existing privacy-preserving methods perform poorly in terms of inference speed, and suffer from the convergence problem during the training stage.
The drastic increase in language models' parameters has led to a new trend of deploying models in cloud servers, raising growing concerns about private inference for Transformer-based models.
% Concerns over the private inference of NLP services and Transformer models have grown in recent years. 
Existing two-party privacy-preserving techniques, however, only take into account natural language understanding (NLU) scenarios.
Private inference in natural language generation (NLG), crucial for applications like translation and code completion, remains underexplored.
% , leaving open the problem of private inference for text generation, involving practical applications including translation, dialogue, code completion, and so on. 
In addition, previous privacy-preserving techniques suffer from convergence issues during model training and exhibit poor inference speed when used with NLG models due to the neglect of time-consuming operations in auto-regressive generations. To address these issues, we propose a fast private text generation framework for Transformer-based language models, namely MERGE.
MERGE reuses the output hidden state as the word embedding to bypass the embedding computation and reorganize the linear operations in the Transformer module to accelerate the forward procedure. 
%Based on these two optimizations, 
Extensive experiments show that MERGE achieves a 26.5x speedup to the vanilla encrypted model under the sequence length 512, and reduces 80\% communication cost, with an up to 10x speedup to state-of-the-art approximated models.
\end{abstract}

\section{Introduction}

Recently, from pre-trained language models (PLMs) to large language models (LLMs), Transformer~\cite{transformer} based models have attracted significant attention because of their exceptional performance in downstream tasks. 
Due to the high demand for computing power, this growth of model parameters also has caused the trend of hosting models to cloud service providers, which raises wide concerns about privacy inference and training. 
For example, existing natural language processing (NLP) services like Copilot\footnote{\url{https://github.com/features/copilot}} and ChatGPT\footnote{\url{https://chat.openai.com}} require users to submit their queries in plain text, which may contain confidential information such as source code, medical information, and personal preferences. 

% \footnote{\url{https://www.newsdirectory3.com/less-than-20-days-after-samsung-introduced-chatgpt-an-accident-occurred-semiconductor-secrets-may-have-been-leaked-xfastest-news/}}

% Moreover, sometimes the weights of service models cannot be accessed by users because they are confidential and expensive to train.
 % ($\sim$100x slower in BERT and $\sim$1000x slower in GPT-2)
 
To alleviate the privacy problem,  recent works~\cite{iron,thex} have developed two-party secure inference services for PLMs by secure Multi-Party Computation (MPC). MPC ensures privacy by encrypting user data and model weights as well as sharing them secretly. However, PLMs inference under MPC is considerably slow compared to the plain-text version, which limits its application in real-world services. To address this issue, several works have attempted to simplify the bottleneck operations such as activation functions and softmax in the Transformer model. For instance, \cite{delphi} uses Neural Architecture Search (NAS) to replace the activation functions with linear layers, and \cite{mpcformer} approximates the exponential operation with polynomial functions.

% \begin{figure}[t]
% \centerline{\includegraphics[width=\linewidth]{figs/introductions.drawio.pdf}}
% \caption{\text{MPC-based Private Inference of Language Models.}}{\label{fig:intro}}
% \end{figure}

\begin{figure*}[t]
\includegraphics[width=0.95\linewidth]{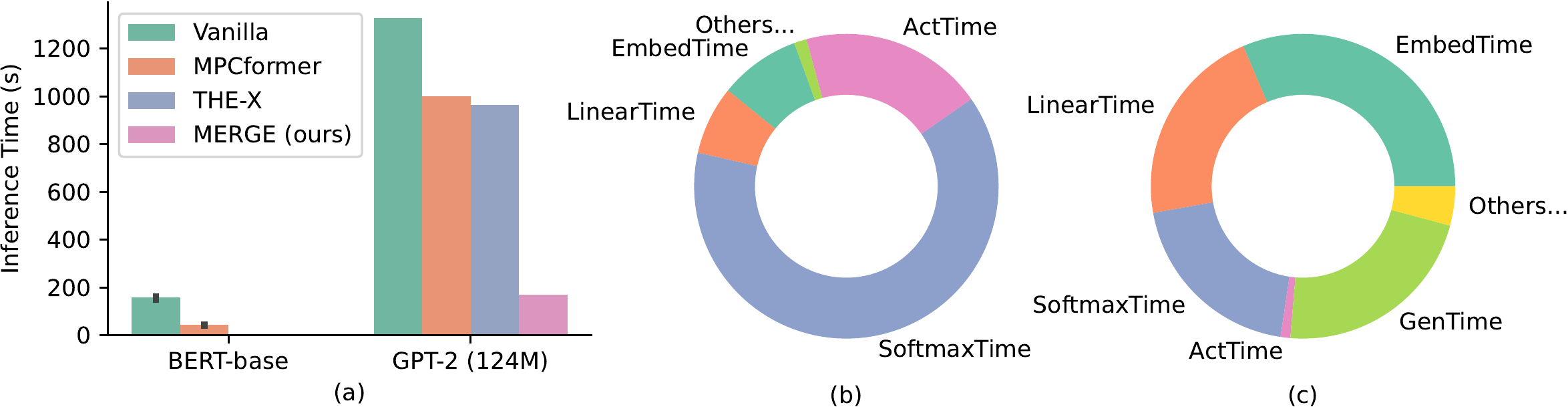}
\caption{
\textbf{Encrypted inference time comparisons among BERT-base and GPT-2 with sequence length 128}, where MPCformer, THE-X, and MERGE in Fig. (a) are different private inference methods. While nonlinear operations such as softmax (SoftmaxTime) and activation functions (ActTime) account for a substantial portion of inference time in BERT-base (Fig. (b)), the inference time is more balanced across operations for NLG models (Fig. (c)), with non-trivial time consumption from linear computations (LinearTime), embedding table query (EmbedTime), and token sampling (GenTime).}
% While existing methods (e.g. MPCformer) achieved significant improvements in NLU, the speedup of NLG is poor due to the extra generation time and auto-regressive forward inferences.}
\label{fig:comparison}
\end{figure*}

Though designed for Transformer, existing works~\cite{iron,thex,mpcformer} solely explore the scenario of natural language understanding (NLU) (e.g., on the GLUE~\cite{glue} benchmark).
Unfortunately,  we observe that they have no significant improvements in natural language generation (NLG) tasks (cf., Fig. \ref{fig:comparison}). 
By illustrating the bottleneck of NLU and NLG inference procedures, we find that auto-regressive generation used in PLMs suffers from extra time cost in \emph{embedding table query} and \emph{token sampling} (i.e., GenTime), which slows down the whole inference procedure heavily.

In this paper, we explore accelerating the generation procedure of language models. Different from existing works that merely approximate the nonlinear operations, we consider the optimization at the architecture level and attempt to reorganize and simplify the whole generation procedure as well as Transformer modules.
To this end, we propose \emph{MERGE} (short for \textbf{M}PC-based \textbf{E}mbedding \textbf{R}esending \textbf{GE}neration), a fast and easy-to-adopt framework for private text generation. \emph{MERGE} is compatible with previous MPC-based works (e.g., MPCformer, THE-X, and IRON) and mainstream PLMs  (e.g., GPT-2~\cite{gpt2}, T5~\cite{t5}, and Bart~\cite{bart}). In concrete, \emph{MERGE} can simplify the time-consuming operations in NLG such as embedding query and token sampling. To achieve that, we first put forward a strategy called \textbf{embedding resending}, which directly uses the output hidden state as the new input token embedding.
Embedding resending helps to bypass the \emph{embedding table query} operation and decouple the computation between \emph{forward representation learning} and \emph{next token sampling}.
Besides, following the recent research~\cite{ca} in attention mechanism, we approximate \emph{self-attention} with \emph{constant attention} matrices and merge tensor computations in the Transformer module before inference.
Nevertheless, these two strategies are challenging because: 1) PLMs are usually sensitive to input embeddings, while there are some unavoidable errors in the generated embeddings; 2) constant attention in our \textit{merge module} might hurt the performance of PLMs.
To address the above challenges, we first propose an embedding alignment and augmentation task to enhance the robustness of PLMs about input embeddings.
Besides, we employ a weighted distillation training task for approximation models, which allows us to alleviate the negative effects of constant attention.
Our empirical experiments on popular text generation tasks such as E2E~\cite{e2e}, Multiwoz 2.1~\cite{multiwoz21}, and DailyDialog~\cite{dailydialog} demonstrate the effectiveness of \emph{MERGE}. 
Specifically, it achieves a considerable speedup of 7.75x to GPT-2 and 10.89x to T5 under the sequence length 128, and 26.5x under sequence length 512, while maintaining an acceptable performance with losses in BERTscore~\cite{bertscore}, BARTscore~\cite{bartscore}, and Rouge-L~\cite{rouge} of only 0.02 (under 0.92), 0.14 (under -2.90), and 0.03 (under 0.44), respectively. Source code of experiments can be found here: \url{https://github.com/liangzid/MERGE}.

\section{Related Work}
\label{sec:related}

% \ref{sec:XXX}.
%\noindent
%\textbf{Approximations in MPC Systems.} 
Although existing MPC techniques can provide secure inference for neural networks, they usually suffer from prohibitively high communication delays and computation costs. 
This is primarily due to the critical nonlinear operations within neural networks. 
Therefore, some works aim to approximate these bottleneck operations in neural networks. 
For instance, \cite{thex} replaces the GeLU activation function in the Transformer with ReLU, and \cite{iron} reformulates the $Tanh(\cdot)$ function in GeLU based on optimized exponential operations.
Besides, \cite{delphi} approximates the ReLU function with linear layers to replace the MPC method used for ReLU through the garbled circuits with secret sharing and Beaver triples. 
Similarly, \cite{mpcformer} approximates GeLU with ReLU and quadratic functions. 
For the softmax operation in the attention mechanism, \cite{mpcformer} approximates it by $softmax(x)\approx \frac{ReLU(x)}{\sum ReLU(x)}$ or $softmax(x)\approx \frac{(x+c)^{2}}{\sum (x+c)^{2}}$. 

Nevertheless, these approximations were designed for the ``one-time'' inference of NLU models (e.g. BERT), and are not optimized for auto-regressive generative models (e.g. GPT-series) that execute the forward inference multiple times. By contrast, our work focuses on optimizing the overall generation procedure instead of some nonlinear operations, which leads to more transformative performance for Transformer-based language models.
% reorganizes and simplifies the generation procedure as well as the Transformer architecture.

\begin{figure*}[t]
\centering
\includegraphics[width=.9\linewidth]{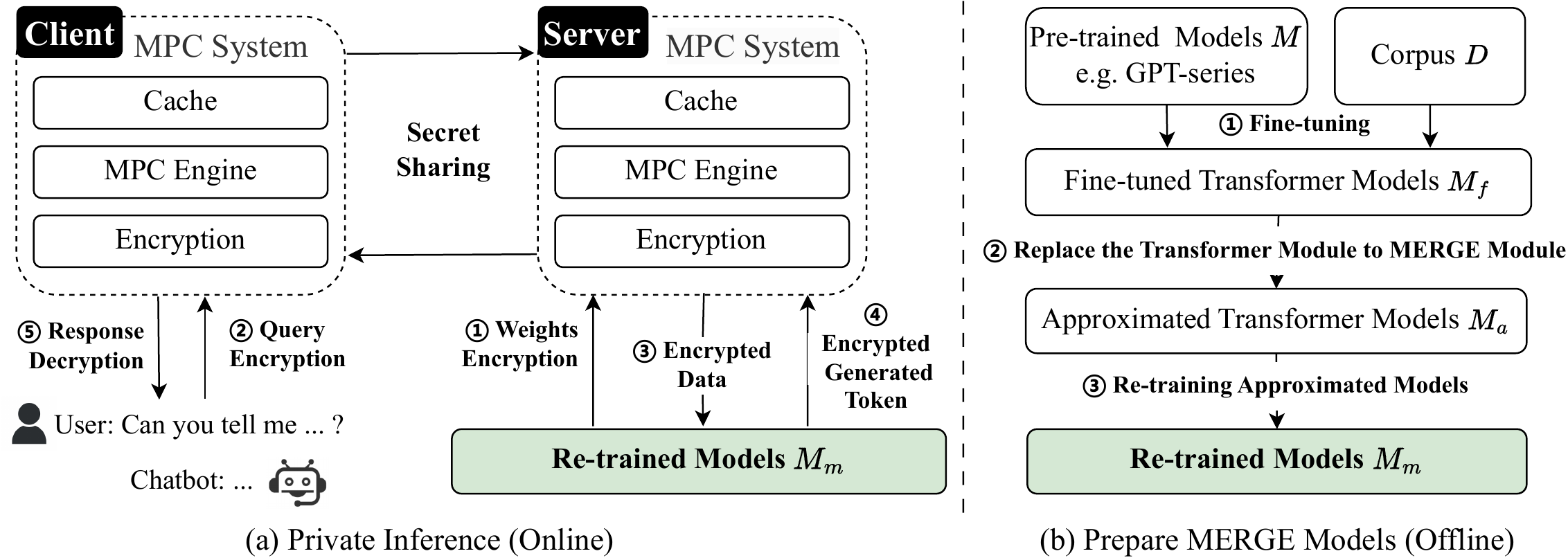}
\caption{\textbf{An overview for \emph{MERGE}}, which achieves the private inference with MPC systems (cf., Fig. (a)) that encrypt the model parameters and the data of a user and connect them by secret sharing and other techniques. As shown in Fig. (b), \emph{MERGE} aims to optimize the deployed model $M_f$ to fit the requirements of MPC systems and accelerate the inference procedure.
}
\label{fig:system}
\end{figure*}

\section{Preliminary}

\subsection{Text Generation with Language Models}\label{sec:ar}
The text generation task (e.g. dialogue) aims to generate the desired sequence $y$ (e.g. the response of the chatbot) under the given prefix text $p$ (e.g. the dialogue context) with the language model $p_{\theta}(y|p)$. 
Typically, existing language models usually generate $y$ in an \emph{auto-regressive} manner, i.e.,

\begin{equation}
\label{eq:1}
p_{\theta}(y|p)=\prod_{t=1}^{N_t}p(x^{y}_{t}|p,x^{y}_{<t}),
\end{equation}
\noindent where $x^{y}_{t}$ denotes the $t$-th generated token of $y$ and $x^{y}_{<t}$ denotes the generated tokens of $y$ at step $t$.

In Eq. (\ref{eq:1}), if we denote the one-hot representation of $(p,x^{y}_{<t})$ as $\textbf{x}_{t}$ with text length $N_{t}$, then the generation procedure consists of  the following three stages:
 
\noindent\textbf{a) Embedding table query:} to obtain the initialized embeddings for each word, i.e., $\textbf{E}_{t}=f_{e}(\textbf{x}_{t})$, where 
$f_{e}(\textbf{x}):\mathbb{R}^{N_{t}\times V}\rightarrow\mathbb{R}^{N_{t}\times d}$ is the embedding layer that maps the $V$-length index representation into the $d$-dimension semantic space to obtain the semantic embeddings $\textbf{E}_t$.

\noindent\textbf{b) Representation learning:} to obtain the representations of inputs considering the contexts, i.e., $\textbf{h}_{t}^{n_{l}}=\textbf{f}_{tr}(\textbf{E}_{t}')$, where $\textbf{f}_{tr}:\mathbb{R}^{N_{t}\times d}\rightarrow\mathbb{R}^{N_{t}\times d}$ is an $n_{l}$-layer transformer model, $\textbf{h}_{t}^{n_{l}}$ is the output hidden state, and $\textbf{E}_{t}'$ is the combination of positional embeddings, token embeddings $\textbf{E}_{t}$, and others.

\noindent\textbf{c) Next token sampling} that generated the new token, i.e., $x^{y}_{t}\sim f_{cls}(\textbf{h}_{t}^{n_{l}})[N_{t}]$, where $f_{cls}(\textbf{h}^{n_{l}}_{t}):\mathbb{R}^{N_{t}\times d}\rightarrow\mathbb{R}^{N_{t}\times V}$ is the linear head for token prediction, and $f_{cls}(\textbf{h}_{t}^{n_{l}})[N_{t}]$ which is the $N_{t}$-th element of $f_{cls}(\textbf{h}_{t}^{n_{l}})$ denotes the probability distribution of in vocabulary for current sampled token, and $\sim$ denotes the sampling strategy (e.g., greedy search) to obtain the new generated token $x_t^y$ according to vocabulary distribution $f_{cls}(\textbf{h}_t^{n_l})[N_t]$.

%, {\color{red} recording the probability distribution of taking each token in the vocabulary as output}.

\subsection{Transformer Module}\label{sec:tr}

In the above representation learning step, the Transformer model $\textbf{f}_{tr}$ can be viewed as a stack of transformer modules. 
Here, we introduce the three key components of transformer module $f_{tr}^{n}:\mathbb{R}^{N_{t}\times d}\rightarrow\mathbb{R}^{N_{t}\times d}$ as follows:

\textbf{a) Projection}: to compute the subsequent self-attention, the transformer module first projects the input hidden state to the $(query, key, value)$ tuple, i.e.,
\begin{equation*}
\textbf{Q}^{n},\textbf{K}^{n},\textbf{V}^{n}=W_{Q^n}^{T}\textbf{h}^{n-1}, W_{K^n}^{T}\textbf{h}^{n-1},W_{V^n}^{T}\textbf{h}^{n-1},    
\end{equation*}
where $W_{Q^n},W_{K^n},W_{V^n} \in \mathbb{R}^{d\times (d/N_{h})\times N_{h}}$ are $N_{h}$-head projection matrices. 
Particularly, we have $\textbf{h}^{0}=\textbf{E}'_{t}$.
% where $E'_{t}$ is the aggregated embeddings of positional embedding, token embedding $E_{t}$, and others.

\textbf{b) Self-Attention}
% \footnote{Noted that there are some slight differences for cross attention, e.g. in cross attention $\textbf{K}$ and $\textbf{V}$ are calculated with the output hidden state of the encoder. While it has no impact on our method MERGE, we will simply discuss the situation of self-attention.}
is proposed to aggregate the context information into a new representation for each word base on the above $(\textbf{Q}^{n},\textbf{K}^{n},\textbf{V}^{n})$ tuple.
Firstly, it calculates the correlation between contextual words, i.e., $A^n=f_{dr}(softmax(\textbf{Q}^{n}\cdot({\textbf{K}^{n}})^{T}/\sqrt{d_{k}}))$, where $A^{n}\in \mathbb{R}^{N_{h}\times N_{t}\times N_{t}}$ denotes the $N_{h}$-head attention matrix and $d_{k}=d/N_{h}$.
Then, the new representations are weighted aggregated based on the above attention matrix, i.e., $\textbf{x}^{n}_{att}=f_{ln}(f_{dr}(W_{d^{n}}^{T}\cdot(\text{Concat}(A^{n}\cdot \textbf{V}^{n}))+b_{d^{n}})+\textbf{h}^{n-1})$, where $W_{d^{n}}\in \mathbb{R}^{d\times d}$ is the weight matrix, $b_{d^{n}}\in \mathbb{R}^{d}$ is the bias, f$_{dr}$ denotes the dropout operation~\cite{dropout}, and $f_{ln}$ is the layer normalization~\cite{lyn},
\begin{equation}
\label{eq:3}
f_{ln}(\textbf{x})={\frac{\textbf{x}-E[\textbf{x}]}{\sqrt{Var[\textbf{x}]}+\epsilon}\odot\gamma+\beta},
\end{equation}
in which $\epsilon$ is a tiny number, $\odot$ denotes the element-wise product, and $E[\textbf{x}]$ and $Var[\textbf{x}]$ denote the mean and variance of $\textbf{x}$, respectively.

\textbf{c) Feed forward:} to compute the output hidden state, a two-layer MLP is used, i.e., $\textbf{h}^{n}=f_{ln}(f_{dr}(W_{O}^{n T}\cdot (\text{Act}(W_{I}^{n T}\cdot \textbf{x}^{n}_{att}+b_{I}^{n})+b_{O}^{n})+\textbf{x}^{n}_{att})$, where $W_{I}^{n}\in\mathbb{R}^{d \times d_{I}}$  and $W_{O}^{n}\in\mathbb{R}^{d_{I} \times d}$ are weighted matrices, $b_{I}^{n}\in  \mathbb{R}^{d_{I}}$ and $b_{O}^{n}\in \mathbb{R}^{d}$ are bias vectors, $d_{I}$ is the dimension of intermediate states, and $\text{Act}(\cdot )$ denotes the activation functions such as ReLU~\cite{relu} or GeLU~\cite{gelu}.

% \subsection{Approximations}
% To approximate the non-linear operations in Section \ref{tr}, existing MPC-based models have proposed several approximations. For the activation function $GeLU$, THE-X and MPCformer 

\begin{figure*}[t]
\centerline{\includegraphics[width=.9\linewidth]{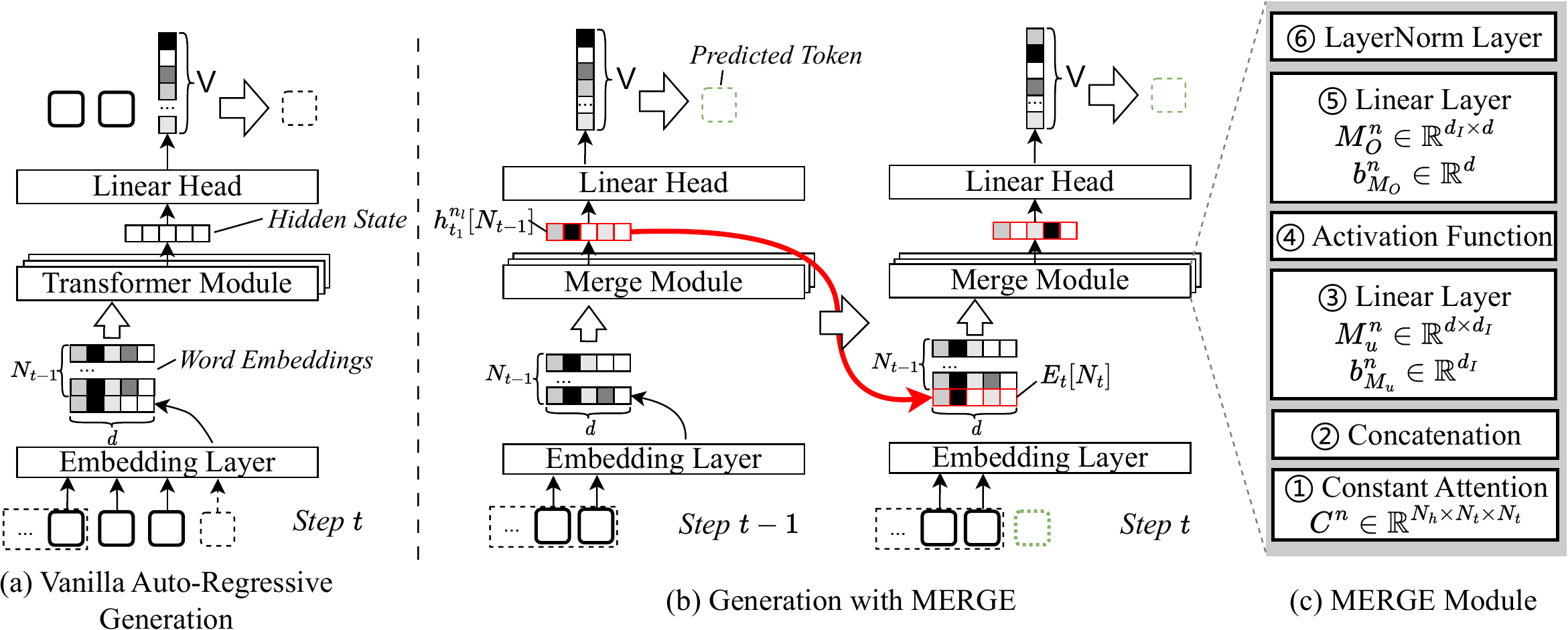}}
\caption{
\textbf{Generation procedure and model architecture of \emph{MERGE}.}
Compared to vanilla auto-regressive generation (cf., Fig. (a)), ER (cf., Fig. (b)) reuses the representation of Transformer model as word embedding of the token input, and thus avoids frequent calls to the embedding layer. Besides, MM (cf., Fig. (c)) replaces dynamic attention with constant attention and merges the Transformer module into six operations to reduce the time costs in softmax and linear computations.
}
\label{fig:ourmethod}
\end{figure*}

\section{Our Method}\label{sec:merge}

\subsection{Overview}

% {\color{red} Fig. \ref{fig:system} shows the private text generation framework of the MPC system and the overview of our \emph{MERGE} method.
% Since MPC methods are sensitive to computation, inference of the base Transformer model in MPC is slow.
% The goal of \emph{MERGE} is to XXXXXX.
% To achieve this goal, we propose A and B.}

Fig. \ref{fig:system} gives an overview of how MPC systems work in private text generation and the role \emph{MERGE} plays in the whole framework. As we can see, to enable private computations among multiple parties, the MPC system encrypts both the texts and the model parameters, and then send them securely with various secure techniques.
% , including garbled circuits~\cite{yao,gol}, fully homomorphic encryption (FHE)~\cite{gen}, and homomorphic secret sharing (HSS)~\cite{boy}, to obtain an encrypted generation results for users. Thanks to the rich support of existing MPC methods, it is practicable to implement the private inference of Transformer models. Therefore, rather than building a new MPC system, this paper focuses on accelerating the private generation procedure for Transformer-based language models.
We detail the MPC system used in this paper in Appendix A. Since it is imperative to accelerate the inference (shown in Fig. \ref{fig:comparison}), our method \emph{MERGE} aims to export an optimized model $M_m$ to the original model $M_f$ for MPC systems, which involves a two-step re-training procedure. Especially, it consists of an approximation of Transformer architecture, called \emph{merge module} (MM), and a training task for a new generation strategy, called \emph{embedding resending} (ER). Fig. \ref{fig:ourmethod} illustrates the details of them.

\subsection{Embedding Resending (ER)}
% {\color{red} To speed up the generation process, we propose a novel ER strategy, whose core idea is XXXX to avoid the time-consuming operation and xxx to decouple XXX.}

As shown in Fig. \ref{fig:ourmethod}(b), the core idea of ER is borrowing the representation of Merge Module as the word embedding (shown as the red line).
In detail, ER regards the hidden state ($\textbf{h}_{t-1}^{n_{l}}[N_{t-1}]$) at step $t-1$ as $\textbf{E}_{t}[N_{t}]$ at step $t$, i.e.,

% strategy aims to speed up the generation process by avoiding time-consuming operations (e.g., \emph{embedding table query}) and decoupling the computation between \emph{representation learning} and \emph{token sampling}. 

\begin{equation}
\label{eq:er}
\textbf{E}_{t}=[\textbf{E}_{t-1};\textbf{h}_{t-1}^{n_{l}}[N_{t-1}]]=[\textbf{E}_{0};\textbf{h}_{t-1}^{n_{l}}],
\end{equation}
where $\textbf{E}_{0}$ denotes the token embeddings of the prefix $p$ and ``$;$'' denotes the concatenation operation.  
In this way, our ER strategy achieves two aspects of optimization: 1) since we obtain $\textbf{E}_{t}[N_{t}]$ without sending new token into the Embedding Layer, we avoid the time-consuming \emph{embedding table query}; 2) since we only require the hidden states instead of sampled tokens in following generation procedure, we can execute the \emph{token sampling} and \emph{representation learning} in parallel.
% When we get the sequence of output hidden state $[h_1, \dots, h_{N_t}]$, we can call the token sampling in parallel to get output token sequence $[x_1, \dots, x_{N_t}]$.

% \noindent
% \textbf{Representation Alignment}
\noindent \textbf{Alignment optimization.} Eq. (\ref{eq:er}) assumes \emph{embedding table query} as the inverse procedure of \emph{next token sampling}, which implies that hidden states and token embeddings are in the same representation space.
% , and the embedding layer $f_{e}$ is the inverse function of $f_{cls}$. 
Therefore, to align the representations $\textbf{h}^{n_{l}}_{t-1}[N_{t-1}]$ and $\textbf{E}_{t}[N_{t}]$, we design a training task that maximizes the cosine similarity between these vectors, i.e.,
\begin{equation}\small
\label{eq:cos}
\mathcal{L}_{c}=\frac{1}{N_{tr}\cdot N}\sum_{i}^{N_{tr}}\sum_{t=1}^{N}{1-cos(\textbf{h}^{n_{l}}_{i,t-1}[N_{t-1}], \textbf{E}_{i,t}[N_{t}])},
\end{equation}
where function $cos$ computes the cosine of the angle between two vectors, $N_{tr}$ is the number of train set, and $N$ denotes the sequence length. Here we select the cosine similarity instead of mean square error (MSE) because the inner product (e.g., \emph{self-attention}) plays a key role in the Transformer module.

\noindent \textbf{Robustness Optimization.}
Besides, we observe that the error of token embeddings significantly impacts the performance of the Transformer model $\textbf{f}_{tr}$ and leads to nonsensical sentence generation with the MSE value over $0.05$ (cf., Fig. \ref{fig:noise}). 
To enhance the robustness of $\textbf{f}_{tr}$, we introduce an \emph{embedding augmentation} method that first masks each element $e_{t}$ in $\textbf{E}_{t}$ with a probability $p$, and then adds a uniform noise sampled from a small interval $(-\epsilon,\epsilon)$ i.e.,
\begin{equation}
\label{eq:4}
\tilde{e}_{t}=m_{t}\cdot(e_{t}+n_{t}),
\end{equation}
where $m_{t}\sim \text{Bernoulli}(1-p)$ and $n_{t}\sim \text{Uniform}(-\epsilon,\epsilon)$. Based on the noised token embeddings $\Tilde{\textbf{E}}_t$, the cross-entropy loss can be reformulated as, 
\begin{equation}
\label{eq:2}
\mathcal{L}_{ce}=\frac{1}{N_{tr}\cdot N}\sum_{i}^{N_{tr}}\sum_{t=1}^{N}\textbf{x}_{t}[N_{t}]\cdot\text{log}f_{cls}(\textbf{f}_{tr}(\tilde{\textbf{E}}_{t}))[N_{t}].
\end{equation}
In this way, for a word embedding input in a noisy range, $\textbf{f}_{tr}$ will learn to obtain a similar hidden state.

Therefore, the overall train loss is formulated as,

\begin{equation}
\label{eq:5}
\mathcal{L}=\lambda\mathcal{L}_{c}+(1-\lambda)\mathcal{L}_{ce},
\end{equation}
where $\lambda \in [0,1]$ is a weighting factor.

% \usepackage{multirow}
% \usepackage{graphicx}
\begin{table*}[t]
\centering
\resizebox{0.7\textwidth}{!}{%
\begin{tabular}{lrrrrr}
\Xhline{2.1pt}
 & \multicolumn{3}{c}{Time/ Communication Time (s)} & \multicolumn{1}{l}{} & \multicolumn{1}{r}{} \\ \cline{2-4}
\multirow{-2}{*}{Model} & \multicolumn{1}{c}{Embed} & \multicolumn{1}{c}{Linear} & \multicolumn{1}{c}{Softmax} & \multicolumn{1}{r}{\multirow{-2}{*}{Total Time (s)}} & \multicolumn{1}{r}{\multirow{-2}{*}{Speedup}} \\ \hline
\multicolumn{6}{c}{\emph{GPT2-base (124M)}} \\ \hline
CrypTen & 321.44/52.33 & 251.93/\underline{74.21} & 454.61/113.96 & 1328.26 & 1x \\
MPCformer (sm2relu) & 316.75/51.55 & 253.57/76.56 & 181.14/45.59 & 1001.41 & 1.33x \\
MPCformer (sm2quad) & 318.16/50.88 & 253.30/75.16 & 152.45/37.40 & 972.50 & 1.36x \\
THE-X & 329.29/58.30 & 258.00/80.21 & 87.71/19.28 & 965.79 & 1.37x \\
MERGE (ER+MM) & \textbf{5.17}/\textbf{0.87} & \textbf{157.50}/\textbf{53.97} & \textsuperscript{$\dagger$}\textbf{0.00}/\textbf{0.00} & \textbf{171.38} & \textbf{7.75x} \\
MERGE (only ER) & \underline{5.41}/\underline{0.95} & 260.36/80.00 & 477.76/124.83 & 834.13 & 1.59x \\
MERGE (only MM) & 320.84/50.92 & \underline{250.98}/81.57 & \textsuperscript{$\dagger$}\underline{0.00}/\underline{0.00} & \underline{747.45} & \underline{1.78x} \\ \hline
\multicolumn{6}{c}{\emph{T5 (138M)}}  \\ \hline
CrypTen & 323.46/53.36 & 328.09/96.08 & 693.73/175.57 & 1569.41 & 1x \\
MPCformer (sm2relu) & 327.51/55.36 & 328.61/96.80 & 284.65/75.17 & 1207.63 & 1.30x \\
MPCformer (sm2quad) & 324.81/52.03 & 325.97/92.89 & 235.54/58.47 & 1149.07 & 1.37x \\
THE-X & 316.16/48.58 & 321.90/90.82 & 126.73/25.51 & 1050.28 & 1.49x \\
MERGE (ER+MM) & \textbf{7.62}/\textbf{1.27} & \textbf{131.31}/\textbf{44.11} & \textsuperscript{$\dagger$}\textbf{0.00}/\textbf{0.00} & \textbf{144.02} & \textbf{10.89x} \\
MERGE (only ER) & \underline{8.24}/\underline{1.58} & \underline{211.57}/\underline{65.19} & 596.74/166.50 & 874.36 & 1.79x \\
  MERGE (only MM) & 322.38/51.35 & 221.57/69.22 & \textsuperscript{$\dagger$}\underline{0.00}/\underline{0.00} & \underline{693.30} & \underline{2.26x} \\
  \Xhline{2.1pt}
\end{tabular}%
}
\caption{\textbf{Inference time comparison among different operations} (embedding layer, linear layer, and softmax function), where ER and MM are in our \emph{MERGE} method, and sf2relu and sf2quad approximate \emph{softmax} with \emph{$ReLU(\cdot)$} and quadratic functions, respectively. Results marked with $\dagger$ come from the optimization in MM which replaces the dynamics attention with constant attention, thus having no cost in \emph{softmax}. The remaining tables use the same notation system as outlined here.
% We marked the \textbf{best results} and \underline{secondary best results}.
}
\label{tab:speed}
\end{table*}
% \usepackage{graphicx}
\begin{table*}[h]\centering
\resizebox{0.7\textwidth}{!}{%
\begin{tabular}{lrrrrr}
\Xhline{2.1pt}
Model & \multicolumn{1}{l}{Embed (Byte)} & \multicolumn{1}{l}{Linear (Byte)} & \multicolumn{1}{l}{Softmax (Byte)} & \multicolumn{1}{l}{Total (Byte)} & \multicolumn{1}{l}{Fraction} \\ \hline
\multicolumn{6}{c}{\emph{GPT2-base (124M)}} \\ \hline
CrypTen & 71.41GB & 159.36GB & 1.62GB & 322.54GB& 100.00\% \\
MPCformer (sm2relu) & 71.41GB & 135.54GB & 0.54GB & 317.20GB & 98.34\% \\
MPCformer (sm2quad) & 71.41GB & 135.54GB & 0.07GB & 316.73GB & 98.20\% \\
THE-X & 71.41GB & 135.54GB & 0.50GB & 319.14GB & 98.95\% \\
MERGE (ER+MM) & \textbf{1.15GB} & \textbf{119.89GB} & \textsuperscript{$\dagger$}\textbf{0.00GB} & \textbf{121.76GB} & \textbf{37.75\%} \\
MERGE (only ER) & \underline{1.15GB} & 160.63GB & 1.62GB & \underline{168.51GB} & \underline{52.24\%} \\
MERGE (only MM) & 71.41GB & \underline{119.89GB} & \textsuperscript{$\dagger$}\underline{0.00GB} & 281.88GB & 87.39\% \\ \hline
\multicolumn{6}{c}{\emph{T5 (138M)}} \\ \hline
CrypTen & 147.14GB & 199.97GB & 7.72GB & 380.45GB & 100.00\% \\
MPCformer (sm2relu) & 147.14GB & 199.97GB & 2.73GB &364.74GB  & 95.87\% \\
MPCformer (sm2quad) & 147.14GB & 199.97GB & 0.33GB & 362.33GB  & 95.24\% \\
THE-X & 147.14GB & 199.97GB & 2.97GB & 369.73GB & 97.18\% \\
MERGE (ER+MM) & \textbf{1.73GB} & \textbf{95.66GB} & \textsuperscript{$\dagger$}\textbf{0.00GB} & \textbf{98.03GB}  & \textbf{25.77\%} \\
MERGE (only ER) & \underline{1.73GB} & 120.17GB & 7.56GB & \underline{132.44GB}  & \underline{34.81\%} \\
  MERGE (only MM) & 73.72GB & \underline{95.66GB} & \textsuperscript{$\dagger$}\underline{0.00GB} & 257.89GB & 67.79\% \\
\Xhline{2.1pt}
\end{tabular}%
}
\caption{\textbf{Averaged communication costs of our \emph{MERGE} method for private text generation}, where Fraction represents the percentage of communication volume occupied by each method relative to the vanilla method (CrypTen).}
\label{tab:cb}

\end{table*}

\subsection{The Merge Module (MM)}
To further accelerate the inference process, we also propose the merge module, which is an efficient approximation of the Transformer module, focusing on optimizing the computation of the \emph{linear} and \emph{softmax} functions.

% In this subsection, we focus on designing an efficient approximation of the Transformer module $f_{tr}$, i.e., the merge module $f_{mer}$, to accelerate the inference in the \emph{linear computation} and \emph{softmax} function.

Following recent research~\cite{ca}, we first replace the dynamic self-attention matrix $A^{n}$ with a constant attention matrix $C^{n}\in \mathbb{R}^{N_{h}\times N_{t}\times N_{t}}$. We initialize $C^{n}$ with the average of $A^{n}$ in train set, i.e., 

\begin{equation}
\label{eq:6}
C^{n}=\frac{1}{N_{tr}}\sum_{i}^{N_{tr}}A_{i}^{n}
\end{equation}

Besides, we approximate the layer normalization $f_{ln}$ in \emph{attention} with a simple element-wise multiplication $f_{ln}'(\textbf{x})=\textbf{x}\odot\gamma+\beta$, inspired by the previous work~\cite{thex}. 
Consequently, the attention procedure now can now be approximated as,

\begin{equation}
\label{eq:ca}
\textbf{x}^{n}_{att}=f_{ln}'(f_{dr}(W_{d}^{nT}\cdot(\text{Concat}(C^{n}\cdot \textbf{V}^{n}))+b_{d}^{n})+\textbf{h}^{n-1}).
\end{equation}
% where $f_{ln}'(\textbf{x})=\textbf{x}\odot\gamma+\beta$.

Based on Eq. (\ref{eq:ca}), we simplify the whole computation procedure by reorganizing matrix computations in $f_{tr}$ and merging intermediate linear operations. 
Specifically, we merge the projection operation $W_{V}^{n}$, the linear map $W_{d}^{n}$, the approximated layer normalization function $f_{ln}'$, as well as the first linear map in feed-forward $W_{I}^{n}$ into a single linear layer, i.e., a weighted matrix $M_{u}^{n}\in \mathbb{R}^{d\times d_{I}}$ and a bias term $b_{M_{u}}^{n}\in \mathbb{R}^{d_{I}}$, which are formatted as:

\begin{equation}
  \label{eq:merge}
  \begin{aligned}
M_{u}^{n}&=(W_{V^{n}}\cdot W_{d}^{n}+\textbf{1})\odot \gamma \cdot W_{I}^{n},\\ b_{M_{u}}^{n}&=W_{I}^{nT}\odot\gamma\cdot b_{d}^{n}+W_{I}^{nT}\cdot\beta+b_{I}^{n},
  \end{aligned}
\end{equation}
where $\textbf{1}\in \mathbb{R}^{d\times d}$ is the residual term in attention module. 

As Eq. (\ref{eq:merge}) shows that no parameters dependent on input token embeddings $\textbf{E}_{t}'$, we can pre-compute $M_{u}$ and $b_{M_{u}}$ before the inference stage, thus reducing the computation during model execution. 
Thus, we simplify the entire Transformer module into only three tensor multiplications, i.e.,

\begin{equation}
\label{eq:mm}\small
\begin{aligned}
\textbf{x}_{o}^{n}&=f_{mer}(\textbf{h}^{n-1})\\
&=f_{ln}({W_{O}^{n}}^{T}\cdot\text{Act}({M_{u}^{n}}^{T}\cdot C^{n}\cdot\textbf{h}^{n-1}+b_{M_{u}}^{n})+b_{O}^{n}).    
\end{aligned}
\end{equation}

Although it may appear possible to merge $M_{u}^{n}$ with the previous linear matrix $W_{O}^{n-1}$ in Eq. (\ref{eq:mm}) by approximating the layer normalization $f_{ln}$ with $f_{ln}'$, we choose to keep them separate for the following two reasons. 
Firstly, the merged matrix $W_{O}^{n-1}\cdot M_{u}^{n}\in \mathbb{R}^{d_{I}\times d_{I}}$ has significantly more parameters than $W_{O}$ plus $M_{u}$, since $d_{I}$  is typically larger than $d$. 
Secondly, removing $f_{ln}$ in Eq. (\ref{eq:mm}) will hurt the convergence of the merge module heavily during the training stage. 

Obviously, to derive Eqs. (\ref{eq:merge}) and (\ref{eq:mm}), we need to swap $W_{v}^{n}$ and $C^{n}$, which requires the verification that the matrix multiplications on the tensor $\textbf{h}_{t}^{n-1}$ under different dimensions obeys the commutative law. 
Proof of this assertion is available in Appendix B.
% \ref{sec:proofs}.

% \begin{theorem}[Commutative Multiplication]
%   For tensor $X \in \mathbb{R}^{a\times b}$, $M_{1},M_{2} \in \mathbb{R}^{a\times a}$, and $M_{3} \in \mathbb{R}^{b\times b}$, we have
%   $M_{2}\cdot(M_{3}\cdot(M_{1}\cdotX)^{T})^{T}=cdot(M_{3}\cdot(M_{1}M_{2}\cdotX)^{T})^{T}$

\section{Experiments}
\subsection{Settings}

% \usepackage{graphicx}
\begin{table*}[t]
\centering
\resizebox{0.71\textwidth}{!}{%
\begin{tabular}{lrrrrrr}
\Xhline{2.1pt}
\multicolumn{1}{l|}{Model} & \multicolumn{1}{l}{BERTscore} & \multicolumn{1}{l}{BARTscore} & \multicolumn{1}{l}{NIST} & \multicolumn{1}{l}{Rouge-L} & \multicolumn{1}{l}{METEOR} & \multicolumn{1}{l}{CHRF++}  \\ \Xhline{1.5pt}
\multicolumn{7}{c}{\emph{MultiWoz NLG}~\cite{multiwoz21}} \\ \hline
\multicolumn{1}{l|}{GPT-2 (124M)} & \underline{0.9237} & -2.9020 & 4.7907 & 0.4424 & 0.4900 & 43.2777  \\
\multicolumn{1}{l|}{+ER (no train)} & 0.6860 & -5.0660 & 0.2325 & 0.0707 & 0.0425 & 3.9721   \\ \hline
\multicolumn{1}{l|}{+MPCformer (sf2relu)} & \multicolumn{1}{r}{\textbf{0.9287}} & \multicolumn{1}{r}{\textbf{-2.5377}} & \multicolumn{1}{r}{\textbf{5.7248}} & \multicolumn{1}{r}{\textbf{0.4806}} & \multicolumn{1}{r}{\textbf{0.5792}} & \multicolumn{1}{r}{\textbf{48.8241}}   \\
\multicolumn{1}{l|}{+MPCformer (sf2quad)} & \multicolumn{1}{r}{OOT} & \multicolumn{1}{r}{OOT} & \multicolumn{1}{r}{OOT} & \multicolumn{1}{r}{OOT} & \multicolumn{1}{r}{OOT} & \multicolumn{1}{r}{OOT}   \\
\multicolumn{1}{l|}{+THE-X} & \multicolumn{1}{r}{OOT} & \multicolumn{1}{r}{OOT} & \multicolumn{1}{r}{OOT} & \multicolumn{1}{r}{OOT} & \multicolumn{1}{r}{OOT} & \multicolumn{1}{r}{OOT}   \\ \hline
\multicolumn{1}{l|}{+MERGE (ours)} & 0.8984  & -3.1464  & 3.7444  & 0.3970  & 0.4302 & 36.6983   \\ 
\multicolumn{1}{l|}{+MERGE only ER} & 0.9155 & -2.8057 & {5.0812} & 0.4339 & 0.5102 & 44.2484  \\
\multicolumn{1}{l|}{+MERGE only MM} & {0.9268} & \underline{-2.6277} & \underline{5.6524} & \underline{0.4778} & \underline{0.5647} & \underline{47.7262}  \\

  \Xhline{1.5pt}
\multicolumn{7}{c}{\emph{CommonGen}~\cite{commongen}} \\ 
\hline
\multicolumn{1}{l|}{GPT-2 (124M)} & \textbf{0.9336} & \textbf{-3.4710} & \textbf{3.7840} & \textbf{0.2744} & \textbf{0.3012} & \textbf{27.7038}  \\
\multicolumn{1}{l|}{+ER (no train)} & 0.5999 & -4.9864 & 0.0701 & 0.0192 & 0.0066 & 0.9470   \\ \hline
\multicolumn{1}{l|}{+MPCformer (sf2relu)} & \multicolumn{1}{r}{0.8943} & \multicolumn{1}{r}{-4.1436} & \multicolumn{1}{r}{2.1301} & \multicolumn{1}{r}{0.1861} & \multicolumn{1}{r}{\underline{0.2691}} & \multicolumn{1}{r}{\underline{27.6167}}  \\
\multicolumn{1}{l|}{+MPCformer (sf2quad)} & \multicolumn{1}{r}{OOT} & \multicolumn{1}{r}{OOT} & \multicolumn{1}{r}{OOT} & \multicolumn{1}{r}{OOT} & \multicolumn{1}{r}{OOT} & \multicolumn{1}{r}{OOT}  \\
\multicolumn{1}{l|}{+THE-X} & \multicolumn{1}{r}{OOT} & \multicolumn{1}{r}{OOT} & \multicolumn{1}{r}{OOT} & \multicolumn{1}{r}{OOT} & \multicolumn{1}{r}{OOT} & \multicolumn{1}{r}{OOT} \\ \hline

\multicolumn{1}{l|}{+MERGE (ours)} & 0.8821 & -4.2479 & 0.6639 & 0.2025 & 0.1538 & 16.0573  \\
\multicolumn{1}{l|}{+MERGE only ER} & 0.8953 & \underline{-3.8979} & 1.6796 & \underline{0.2430} & 0.2110 & 20.8878   \\
\multicolumn{1}{l|}{+MERGE only MM} & \underline{0.9083} & -4.0885 & \underline{2.2687} & 0.2026 & 0.2058 & 20.9888 \\
% \multicolumn{1}{l|}{no AQ} & - & - & - & - & - & - &  \\

  \Xhline{1.5pt}
\multicolumn{7}{c}{\emph{DailyDialog}~\cite{dailydialog}} \\ 
\hline
\multicolumn{1}{l|}{GPT-2 (124M)} & \textbf{0.8404} & -6.6387 & 0.5429 & 0.1142 & 0.1042 & 11.5089  \\
\multicolumn{1}{l|}{+ER (no train)} & 0.7518 & -6.8820 & 0.1287 & 0.0566 & 0.0526 & 6.8067 \\ \hline
\multicolumn{1}{l|}{+MPCformer (sf2relu)} & \multicolumn{1}{r}{0.8161} & \multicolumn{1}{r}{\underline{-6.3494}} & \multicolumn{1}{r}{\textbf{1.1102}} & \multicolumn{1}{r}{\underline{0.1322}} & \multicolumn{1}{r}{\underline{0.1261}} & \multicolumn{1}{r}{\underline{12.0713}} \\
\multicolumn{1}{l|}{+MPCformer (sf2quad)} & \multicolumn{1}{r}{OOT} & \multicolumn{1}{r}{OOT} & \multicolumn{1}{r}{OOT} & \multicolumn{1}{r}{OOT} & \multicolumn{1}{r}{OOT} & \multicolumn{1}{r}{OOT} \\
\multicolumn{1}{l|}{+THE-X} & \multicolumn{1}{r}{OOT} & \multicolumn{1}{r}{OOT} & \multicolumn{1}{r}{OOT} & \multicolumn{1}{r}{OOT} & \multicolumn{1}{r}{OOT} & \multicolumn{1}{r}{OOT} \\ \hline

\multicolumn{1}{l|}{+MERGE (ours)} & 0.8213 & \textbf{-6.2384} & 0.3674 & 0.1233 & 0.0955 & 7.8091 \\
\multicolumn{1}{l|}{+MERGE only ER} & 0.8205  & -6.5515 & 0.1069 & {0.1301} & 0.0833 & 6.5819 \\
\multicolumn{1}{l|}{+MERGE only MM} & \underline{0.8343}  & -6.5800  & \underline{1.0499}  & \textbf{0.1525} & \textbf{0.1364} & \textbf{14.9039} \\
% \multicolumn{1}{l|}{no AQ} &  &  &  &  &  &  &  \\

  \Xhline{2.1pt}
\end{tabular}%
}
\caption{Performance experiments of our \emph{MERGE} method for private text generation.}
\label{tab:res}
\end{table*}

\noindent
\textbf{Datasets.}
We evaluate \emph{MERGE} on three representative text generation tasks, including Multiwoz~\cite{multiwoz21}, a human-human multi-turn task-oriented dialogue corpus, DailyDialog~\cite{dailydialog}, a multi-turn chitchat dataset, and CommonGen~\cite{commongen}, a hard-constrained controlled text generation benchmark. 
% The details of datasets can be seen in Appendix C.1.
% \ref{app:datasets}.

\noindent
\textbf{Baselines.}
We compare \emph{MERGE} with state-of-the-art private inference models and frameworks, including \text{THE-X}~\cite{thex}, one of the first approximation architecture of Transformer models, \text{MPCformer}~\cite{mpcformer}, the approximated model that aims to accelerate the inference procedure of Transformer, and~\text{Crypten}
% \footnote{https://github.com/facebookresearch/CrypTen}~\cite{crypten}
, one PyTorch version of the general MPC implementations based on secret sharing.

% \item Private Transformer~\cite{crypten}, one of the crypten implementation of Transformer.

\noindent
\textbf{Evaluation Metrics.}
We evaluate \emph{MERGE} in two dimensions: inference speed, and the effectiveness of approximation models. For inference speed, we record both the computation time and the communication volume for each method. For the effectiveness of PLMs, we use Meteor~\cite{meteor}, CHRF++~\cite{chrf}, NIST~\cite{nist}, ROUGE family~\cite{rouge}, BERTscore~\cite{bertscore}, and BARTscore~\cite{bartscore} as the metrics.
% A detailed introduction can be found in Appendix C.2.
% \label{app:metrics}.

\subsection{Implementation Details}
We use GPT-2 (124M)~\cite{gpt2} as the basic evaluation backbone, with max sequence length $128$. We trained all models under the learning rate $3\times 10^{-5}$, batch size $4$ with $3$ epochs, based on the implementation of huggingface Transformers~\cite{transformers}.
As for approximated models, we train our baselines under the same hyperparameter settings in their source code, and train \emph{MERGE} with $50,000$ steps under the learning rate $8\times 10^{-5}$.
All experiments above are on a single 32 GB Nvidia Tesla V100 GPU. Following previous works~\cite{mpcformer}, for the experiments of private inference, we use two 32 GB Nvidia Tesla V100 GPUs to simulate the client and the server, with 10 GbE Ethernet bandwidth. We implement the whole MPC system based on Crypten~\cite{crypten}, a semi-honest MPC framework built on PyTorch.

\subsection{Speed Evaluation}
We evaluate the inference speed under two mainstream NLG architectures, i.e. the pure decoder represented by GPT-2, and the encoder-decoder models represented by T5, to investigate the speedup of ER and MM.
% We evaluate the total inference time of these two architectures as well as the time cost of each operation.
As shown in Table \ref{tab:speed}, \emph{MERGE} obtains a 7.75x speedup to the encrypted GPT-2, and 10.89x to T5, under the sequence length 128, while existing methods merely give the speedup less than 2x. 
Besides, the ER strategy obtains a 59x speedup on Embed Time, which saves 98.3\% inference time in \emph{embedding table query}. 
% Regarding Linear Time, it decouples the linear head $f_cls$ in Sec. \ref{sec:tr} \textbf{c)}, thus reducing 
Different from ER, MM merges the softmax into the results of constant attention, demonstrating a zero cost in softmax, and a slight time decrease in Linear Time.
We also see that MERGE achieves a higher speedup on T5 than GPT-2, which might be because every self-attention module of T5 follows with a cross-attention module possessing a much higher time proportion on linear computations and softmax. 

Under the same settings of Table \ref{tab:speed}, we also record the communication cost between the client and the server, shown in Table \ref{tab:cb}. 
In general, Table \ref{tab:cb} reveals a similar phenomenon to Table \ref{tab:speed}. We see that existing methods reduce the communication cost slightly (less than 2\% in GPT-2), while our method reduces 62\% communication cost, with 98\% and 25\% on \emph{embedding table query} and \emph{linear operation}, respectively. 
% \begin{figure*}[t]
%     \centering
%     \includegraphics[width=\textwidth]{curve/vary_msl.pdf}
%     \caption{The inference time and communication cost varying generated max sequence lengths.}
%     \label{fig:varySL}
% \end{figure*}
% \begin{figure*}[h]
%     \centering
%     \includegraphics[width=\textwidth]{curve/vary_params.pdf}
%     \caption{The inference time and communication cost varying model parameters.}
%     \label{fig:varyP}
% \end{figure*}
\begin{figure*}[t]
    \centering
    \includegraphics[width=0.93\textwidth]{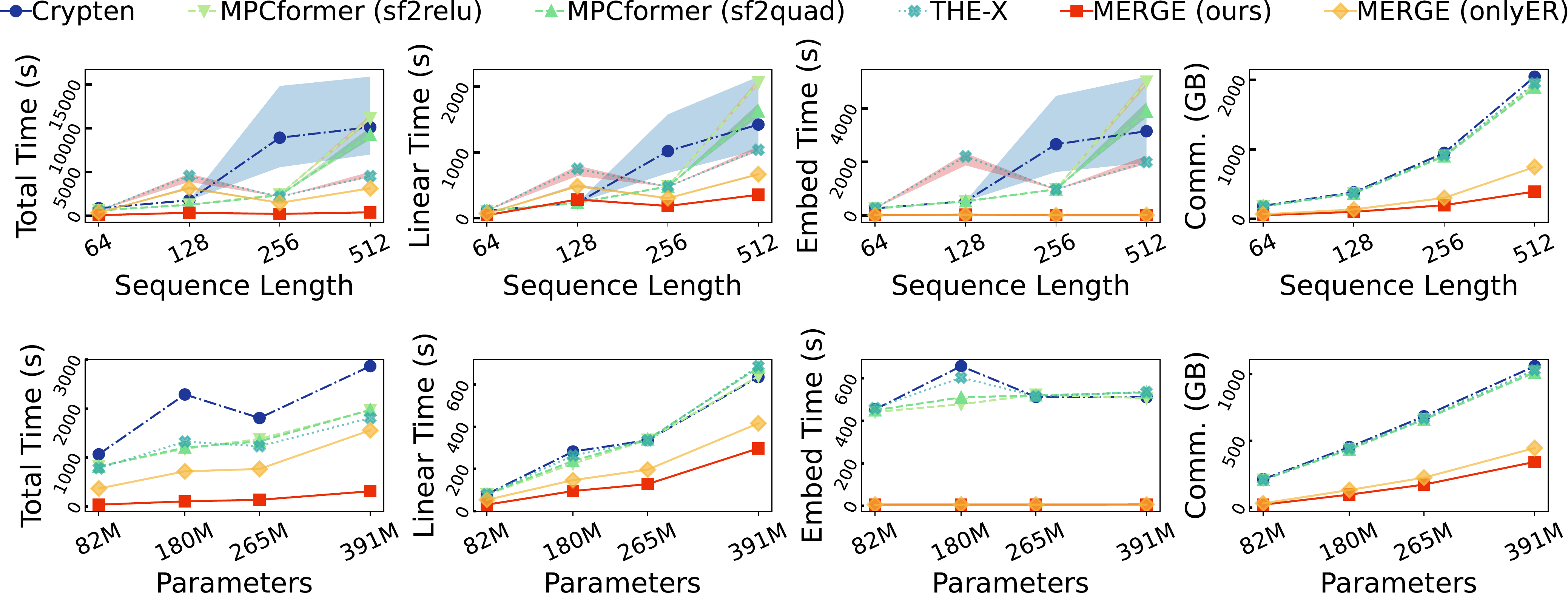}
    \caption{The inference time and communication cost varying generated max sequence lengths and model parameters.}
    \label{fig:varySLP}
\end{figure*}
\begin{figure}[!t]
    \centering
    \includegraphics[width=0.93\linewidth]{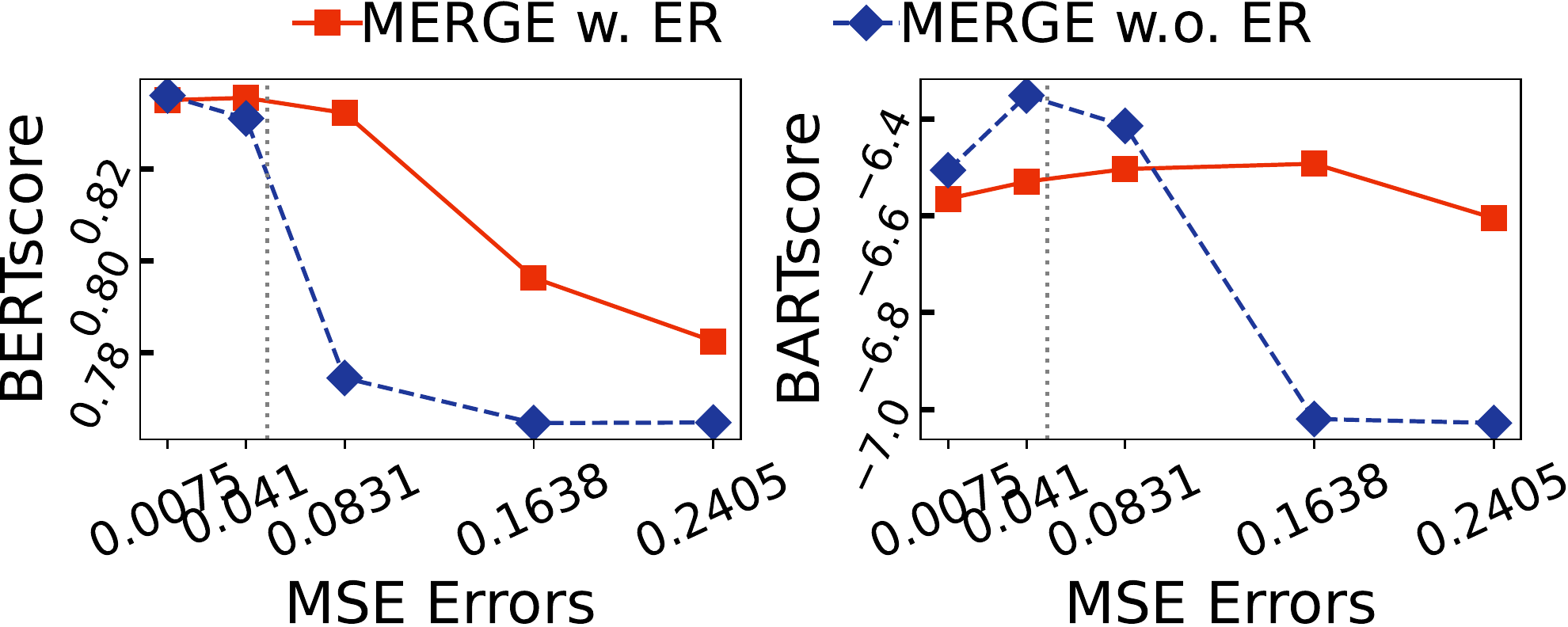}
    \caption{Robustness experiments varying MSE error.}
    \label{fig:noise}
\end{figure}

\subsection{Performance Evaluation}\label{sec:performance}

In addition to inference speed, we also focus on the inference performance between \emph{MERGE} and other MPC frameworks. As shown in Table \ref{tab:res}, our method achieves comparable results to MPCformer and demonstrates strong scores across multiple metrics. For instance, the BERTscore of \emph{MERGE} is lower than MPCformer with ReLU approximation (MPCformer (sf2relu)) by only 0.01, 0.017, and 0.001 in MultiWoz, CommonGen, and DailyDialog, respectively.
Besides, Table \ref{tab:res} indicates that some acceleration methods designed for NLU models are not suitable to text generation models, i.e. they suffer from the convergence problem during training. For instance, THE-X replaces all \emph{layer normalization} operations to the approximate normalization, which we observed will lead to the \textbf{out of time (OOT)} issue. Similarly, the MPCformer that replaces the softmax function with quadratic functions (MPCformer (sf2quad)) faces the same problem, though we train it with an elaborate layer-wise knowledge distillation.
% In addition, we investigate the performance of MERGE among different backbones. As shown in Table \ref{tab:res}, XXXXXXXX.

\section{Analysis}
\noindent
\textbf{Model Size and Sequence Lengths.}
In this section, we dive to explore the effectiveness of our \emph{MERGE} method under longer sequence lengths and larger model parameters.
For sequence length, we set it from 64 to 512 and record the average score as well as the minimum and maximum score for each point. 
Illustrated by Fig. \ref{fig:varySLP} (the upper row), we see that the inference time cost, as well as the communication cost, decreases with the improvements in sequence length. In detail, our \emph{MERGE} method can obtain a 26.5x speedup to the vanilla model and 11.8x to the state-of-the-art model THE-X under sequence length 512, and reduce almost 80\% communication cost. Besides, our embedding resending (ER) strategy can obtain a \textbf{constant} embedding inference time, which is because ER bypasses the \emph{embedding table query}, and thus its embedding time is only related to the generation prefix of samples. 
% \lzm{why mpcformer> crypten?} \lzm{why crypten have larger variance?}

For model parameters, we also evaluate \emph{MERGE} under different model sizes from 82M to 391M and set the sequence length to 128. The parameter experiments in Fig. \ref{fig:varySLP} (the below row) demonstrates that there are no significant improvements in speedup while the model size increases, but our \emph{MERGE} still obtains a significant speedup (\textasciitilde 10x) to existing methods. Besides, our method exhibits a conspicuous positive correlation with the model parameter size in terms of the gap between our method and the baselines, particularly in linear time and the communication cost, which demonstrates the effectiveness of our \emph{MERGE}.

\noindent
\textbf{Robustness of Word Embedding.}\label{sec:robustness}
Illustrated in Fig. \ref{fig:noise}, we add random noise on the embedding of Transformer models and evaluate the decrease of text generation for different generation strategies. In concrete, Fig. \ref{fig:noise} demonstrates that there exists an abrupt decline with MSE error 0.08 for vanilla auto-regressive generation, while our method can resist the decreasing of generation quality.

\section{Conclusion}
To address the problem of private text generation, we propose \emph{MERGE}, a novel framework to accelerate the inference procedure of language models. 
\emph{MERGE} consists of two optimizations, embedding resending and the merge module. The former speeds up the generation by bypassing the embedding query of Transformer models, and the latter optimizes the computation of Transformer modules.
Extensive experiments demonstrate the superiority of \emph{MERGE} both in inference speed and generation quality. 
% In the future, we plan to design a fast and plug-and-play MPC framework for existing language models. 

% \bibliographystyle{plainnat}
% \bibliographystyle{./aaai24.bst}
\bibliography{refs1}

% \newpage

\clearpage
\appendix

\section{Details of the MPC System}
%In this section, we will give a complete overview of our implementation of the MPC system, i.e. the private inference under CrypTen.

\subsection{Preliminary}

\textbf{System Setup}. 
The whole protocol aims to provide a two-party privacy-preserving inference procedure, where there exist two parties, the client and the server. 
The client usually represents the user who owns the inference data, and the server represents the provider of services that owns the language models. 
Under the two-party security setting, our protocol aims to not only ensure that the data of the client cannot be leaked but also to keep the privacy of the model parameters from the server. 

\noindent\textbf{Threat Model}. 
Our protocol assumes both the client and the server are \emph{semi-honest}, i.e., we can ensure both of the parties will follow the protocol and never deviate from it, but each of the parity will not give up the possibility of mining information of the adversary based on the messages it receives from the MPC system.

\noindent\textbf{Private Computation}. 
Our protocol achieves private linear computation based on secret share and Beaver's multiplicative triples. 
Specifically, given an element $x_r$ in the computation procedure, we first transform it to an integer $x \in \mathcal{R}$ under the finite ring $\mathcal{R}$, and then we can split $x$ to $[x]_1$ and $[x]_2$ to guarantee $[x]_1+[x]_2=x$ so that the adversary party only can obtain part of the data of $x$ (e.g., $[x]_2$). 
In this way, the value $x$ under the view of the adversary party is perfectly hidden, and we can use the property of secret sharing to achieve a private addition operation. 
For multiplication, one of the solutions is Beaver's multiplicative triple. 
In concrete, for elements $[x]$ and $[y]$ owned by the client and the server respectively, we can obtain the value of $[x][y]$ privately based on a random element triple $([c],[a],[b])$, where $c=ab$. 
Based on $([c],[a],[b])$, instead of sending $[x]$ or $[y]$ directly, the two parties will send $[\epsilon]=[x]-[a]$ and (or) $[\delta]=[y]-[b]$, and the multiplication of $[x]$ and $[y]$ can be derived as $[x\cdot y]=[c]+\epsilon [b] + [a]\delta +\epsilon\delta$.

\subsection{MPC Computations}

\textbf{Linear Operations} can be implemented by combining the additive secret sharing and Beaver's multiplication triples.

\noindent\textbf{Nolinear Operations} such as softmax and sigmoid in CrypTen are implemented by a two-step process: 1) these operations will be approximated to a linear expression, where CrypTen evaluates exponentials, logarithms, and reciprocals with the limit approximation, the Householder iterations, and the Newton-Rphapson iterations, respectively; 2) the approximated operation can then be implemented by private additions and multiplications.

% \section{Proofs and Computation Details}

\section{The Commutative Law of Tensor Multiplication varying Dimensions}

\newtheorem{theorem}{Theorem}
\begin{theorem}\label{th:cl}
 Given four matrices $\textbf{X} \in \mathbb{R}^{m\times n}$, $A_1 \in \mathbb{R}^{m\times m}$, $B \in \mathbb{R}^{n\times n}$, and $A_2 \in \mathbb{R}^{m\times m}$, let $f(a,b,i)$ denote the matrix multiplication of $a$ and $b$ under the dimension $i$, then we have,
\begin{equation}\label{eq:cl}\scriptsize
f(f(f(\textbf{X}, A_1,1),A_2,1),B,2)=f(f(f(\textbf{X}, A_1,1),B,2),A_2,1).
\end{equation}    
\end{theorem}

\noindent\textbf{Proofs:} Eq \ref{eq:cl} is equivalent to $f(f(\textbf{X},A_2,1),B,2)=f(f(\textbf{X},B,2),A_2,1)$, where
 
\begin{equation}
\begin{aligned}
f(f(\textbf{X},A_2,1),B,2)&=(\textbf{X}^T\cdot A_2)^T \cdot B\\
&=A_2^T\cdot (\textbf{X}\cdot B)\\
&=((\textbf{X}\cdot B)^T\cdot A_2)^T\\
&=f(f(\textbf{X},B,2),A_2,1)\\
\end{aligned}
\end{equation}

% \noindent
We can simplify $f(f(f(\textbf{X},A_1,1),B,2),A_2,1)$ to $f(f(\textbf{X},B,2),A,1)$ based on Theorem \ref{th:cl}, where $A=A_1\cdot A_2$. In this way, we can merge all matrices and biases under the same dimension, shown in Eq. (\ref{eq:merge}).

% \subsection{The Computation Procedure of Equation \ref{eq:mm}}

\section{Concerns \& Limitations}
\subsection{Theoretical comparison}\label{sec: theo}
Suppose the prefix (prompt) length of the text is $N_{p}$, the generated text length is $N_{t}$, the vocabulary size is $V$, the number of Transformer Module is $N_{l}$, the feature and intermediate feature dimension is $d$ and $d_{I}$, respectively, then we have i) the \emph{inference time} of vanilla fine-tuning models and our baselines have the complexity of $\mathcal{O}((N_{t}+N_{p})(Vd+d d_{I} (N_{p}+N_{t}) N_{l})\rightarrow \mathcal{O}(N_{t}Vd+N_{t}^{2}d d_{I}N_{l})$ in \emph{linear operations}, while MERGE reduces it to $\mathcal{O}(N_{p}\cdot Vd+ N_{t}dd_{I}N_{l})\rightarrow\mathcal{O}(Vd+N_{t}d d_{I}N_{l})$; 
 ii) for \emph{communication cost} in \emph{linear operations}, baselines scale as $\mathcal{O}(N_{t}^{2}(N_{l}d d_{I}+Vd))$, whereas MERGE is $\mathcal{O}(N_{t}N_{l}d d_{I})$; 
 iii) Moreover, baselines suffer from an extra cost by approximated \emph{exponential} operations with the complexity of $\mathcal{O}(N_{t}^{2}N_{l})$. The above analysis is supported by the trends shown in Fig.4 of our paper.

\subsection{Converges of MERGE}

\begin{figure}[htbp]
  \centering
\includegraphics[width=0.8\linewidth]{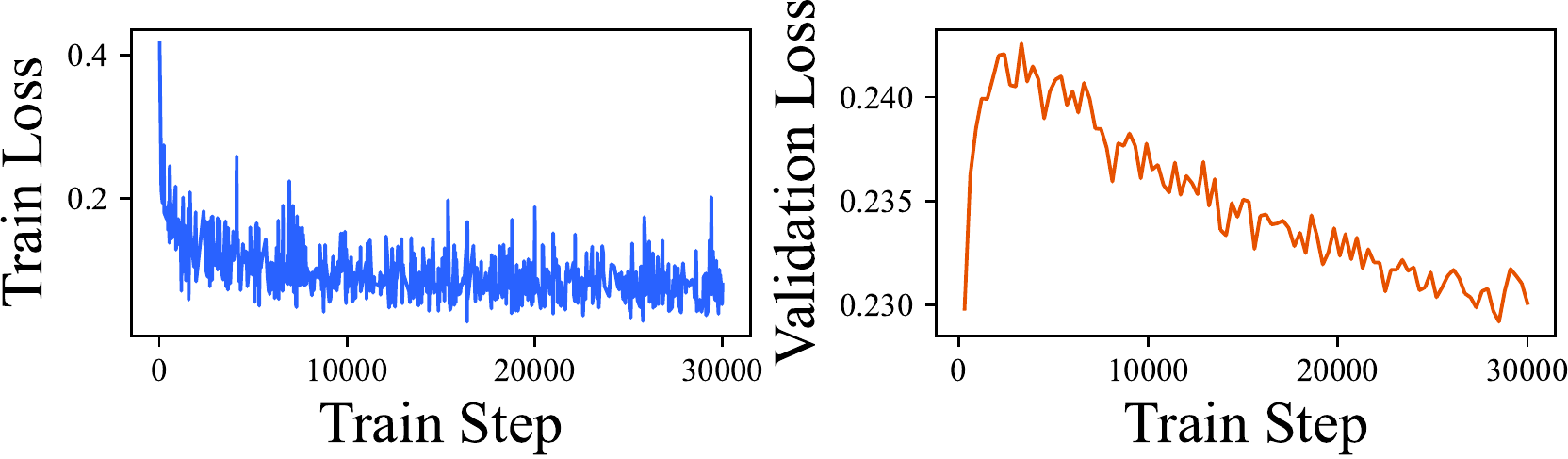}
\caption[]{\text{Loss curves of re-training.}}\label{fig:nothing}
\end{figure}

We re-train models with batch size 16, learning rate 8e-5, and max sequence length 128. Unlike the vanilla fine-tuning procedure, we set the maximum training step as 30,000. We set the dropout rate to 0.6, $\lambda$ to 0.75, and noise to 0.75. It will cost 0.09 hours for every 1,000 steps. Shown in Fig. \ref{fig:nothing}, the train and validation loss converge stably.

The curve of validation loss rises initially, due to the architecture replacement from Transformer to MERGE.

\subsection{Scaleability to Larger Models}
Since Crypten has not supported multi-GPU yet, the maximum parameters we examined for GPT-2 is about 391M. But based on the analysis in Sec. \label{sec:theo} and Fig. 4 in the paper, we estimate that MERGE can reduce almost 50.5\% communication costs for a 20B LLM with sequence length 128, of which the speedup is at least 5.2x. We estimate the speedup will drop because the computation proportion of \emph{attention modules} will rise in LLMs. However, when sequence length exceeds 2,048 (commonly used in LLMs), it will achieve a much better speedup (e.g. up to 10x) to the estimations now.

\end{document}